\documentclass[10pt,twocolumn,letterpaper]{article}

\usepackage[pagenumbers]{cvpr} 

\usepackage{graphicx}
\usepackage{amsmath}
\usepackage{amssymb}
\usepackage{booktabs}
\usepackage{multirow}
\graphicspath{{images/}}

\usepackage[pagebackref,breaklinks,colorlinks]{hyperref}

\usepackage[capitalize]{cleveref}
\crefname{section}{Sec.}{Secs.}
\Crefname{section}{Section}{Sections}
\Crefname{table}{Table}{Tables}
\crefname{table}{Tab.}{Tabs.}
\usepackage{bbding}

\begin{document}

\title{\vspace{-2cm}ViR: the Vision Reservoir}

\author{\textbf{Xian Wei $^\ast $}, \textbf{Bin Wang $^1$}, \textbf{Mingsong Chen $^2$}, \textbf{Ji Yuan $^3$}, \textbf{Hai Lan  $^4$}, \textbf{Jiehuang Shi $^5$}, \\ \textbf{Xuan Tang $^6$}, \textbf{Bo Jin  $^7$}, \textbf{Guozhang Chen $^8$}, \textbf{Dongping Yang $^9$}\\
{\tt\small xian.wei@tum.de} \thanks{Corresponding Author} , {\tt\small 208527042@fzu.edu.cn $^1$}, {\tt\small mschen@sei.ecnu.edu.cn $^2$}, {\tt\small yuanj36@vanke.com $^3$},\\
{\tt\small lanhai09@fjirsm.ac.cn $^4$}, {\tt\small jiehuangs@student.unimelb.edu.au $^5$}, {\tt\small 2265275624@qq.com $^6$},\\
{\tt\small bjin@cs.ecnu.edu.cn $^7$}, {\tt\small ifgovh@163.com $^8$}, {\tt\small yangdempe@gmail.com $^9$}
}

\maketitle

\begin{abstract}

The most recent year has witnessed the success of applying the Vision Transformer (ViT) for image classification.
However, there are still evidences indicating that ViT often suffers following two aspects,
\romannumeral 1) the high computation and the memory burden from applying the multiple Transformer layers for pre-training on a large-scale dataset,
\romannumeral 2) the over-fitting when training on small datasets from scratch.
To address these problems, a novel method, namely, Vision Reservoir computing (ViR), is proposed here for image classification, as a parallel to ViT. 
By splitting each image into a sequence of tokens with a fixed length, 
the ViR constructs a pure reservoir with a nearly fully connected topology to replace the Transformer module in ViT. 
Two kinds of deep ViR models are subsequently proposed  to enhance the network performance.
Comparative experiments between the ViR and the ViT are carried out on several image classification benchmarks.
Without any pre-training process, the ViR outperforms the ViT in terms of both model and computational complexity. Specifically, the number of parameters of the ViR is about 15\% even 5\% of the ViT, and the memory footprint is about $20\% \sim 40\%$ of the ViT. The superiority of the ViR performance is explained by Small-World characteristics, Lyapunov exponents, and memory capacity.
\end{abstract}

\section{Introduction}
\label{sec:intro}

Recently, Vision Transformer (ViT) \cite{dosovitskiy2020image} has been demonstrated remarkable performance across various vision tasks, such as image classification \cite{touvron2021training}, instance segmentation \cite{wang2021end_cvpr}, object detection and tracking \cite{chen2021transformer_cvpr}, with the potential to replace the convolutional neural networks (CNNs) \cite{khan2021transformers}.

While the convolutions in a CNN excel on extracting local features of an image, the ViT is based on the self-attention mechanism that can directly compute global relationships among pixels in various small sections of the image (e.g., $15\times 16$ patches) \cite{li2021localvit,chaudhari2021attentive,correia2021attention,khan2021transformers}. 
Furthermore, compared to CNNs, the ViT assumes minimal prior knowledge about the task at hand, while its performance surpasses CNNs on large-scale datasets with pre-training and fine-tuning, and only requires substantially fewer computational resources to train \cite{dosovitskiy2020image}. 
Its superiority attracts researches to further propose many variants, e.g., DeiT, T2T ViT, PVT \cite{touvron2021training,Yuan2021TokenstoTokenVT,wang2021pyramid}, to handle various downstream tasks.
 \begin{figure}[t]
 \centering
 \includegraphics[scale=0.32]{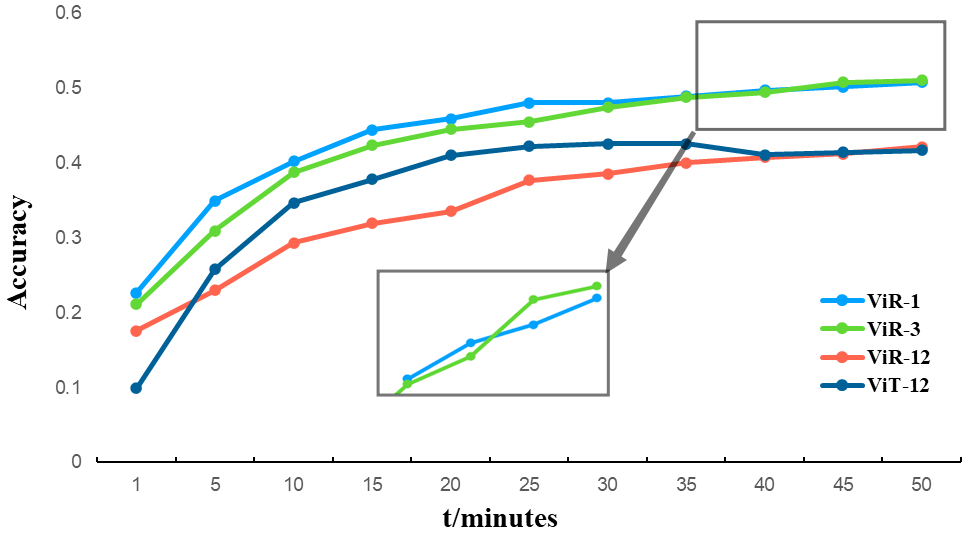}
  \caption{Comparisons of the time consumption by performing the ViR and the ViT on the CIFAR100 dataset. The initial accuracy and the final accuracy of the ViR are higher than the ViT without pre-training. The deep ViR is the parallel structure.
  With the same depth, the time cost of the ViR is much less than the ViT. 
  }
 \label{fig:speed}
  \vspace{-3mm}
\end{figure}

However, pre-training on a large-scale dataset and fine-tuning to downstream tasks are essential for the ViT, which often lead to  the exhaustive and redundant computation, and the memory burden by extra parameters.Additionally, the ViT with multiple Transformer encoder layers often suffers the over-fitting especially when the training data is limited \cite{chen2021visformer_iccv}.

In essence, by treating the image patches as the temporal sequence, the core innovation of the ViT is to use a kernel connection operation, e.g., a dot product, 
to obtain the internal dependencies among image patches, e.g., the spatial and temporal (sequential) coherence between different portions of an image \cite{choromanski2020rethinking_iclr, chen2021transformer_cvpr,khan2021transformers}. 

 \begin{figure*}
 \begin{center}
  \includegraphics[scale=0.275]{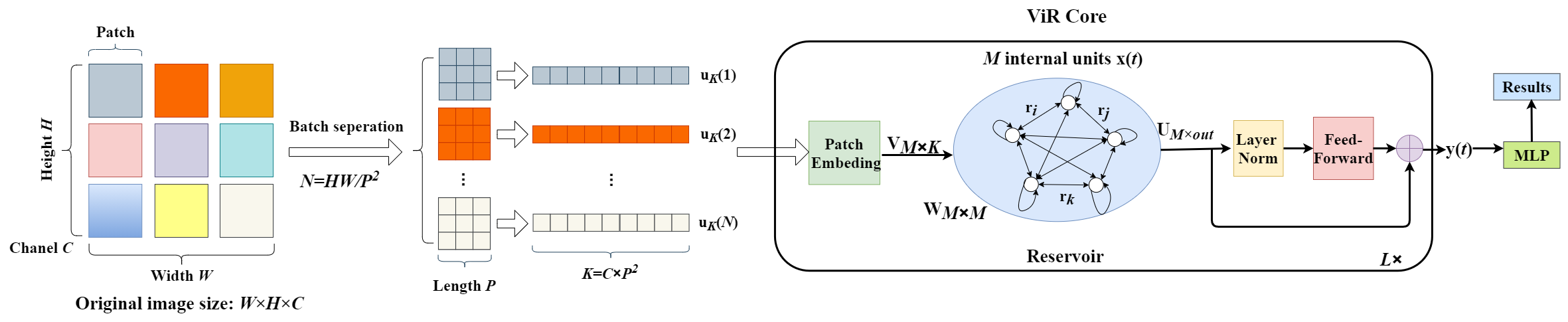}
  \end{center}
  \caption{\textbf{Model overview.} We first split an input image into patches with a proper size, and then flatten each of them into a series of sequence vectors, as the temporal input of the ViR. To have a better performance, the ViR core contains a residual block and could be stacked into a deep structure. }
 \label{fig:vir core}
 \vspace{-3mm}
 \end{figure*}
This motivates us to consider a brain-inspired network, i.e., Reservoir Computing (RC) \cite{lukovsevivcius2009reservoir}, which incorporates intrinsic spatio-temporal dynamics, with much lower consumption on computation and memory, fewer training parameters, and much fewer training samples \cite{jaeger2001echo,maass2002real}.

The main body of RC is a dynamical system, called reservoir \cite{nakajima2021reservoir_book}, which consists of randomly interacting non-linear neurons and fixed weights. 
The non-linearity and disordered connections give rise to the complex internal dynamics of RC. 
Such complex dynamics are speculated to be able to facilitate information processing, with the most benefit at the so-called 'edge of chaos' \cite{legenstein2007edge}. The intrinsic dynamics project inputs to the high-dimensional representation space, facilitating the following regression or classification tasks.
While the recurrent connections in a reservoir are randomly generated and fixed, readout weights are optimized according to the task at hand.
Therefore, RC is a powerful and physically efficient tool in time series related tasks, in terms of the low training cost, fast training speed, few parameters, suitability for hardware, and linear training schemes.

Inspired by the excellent performance of RC on time series tasks, we propose a reservoir mechanism-based model, namely, \textbf{Vision Reservoir computing (ViR)}, to replace the ViT for image tasks.
The ViR follows a similar basic pipeline with the ViT and the whole network is shown in Fig.~\ref{fig:vir core}. 
Without pre-training, the results of the ViR are close to or even better than the ViT on small-scale datasets, but with much fewer parameters and faster training speed, when having the same network depth. Usually, the ViR can achieve considerable performance with layers fewer than the number of encoders of the ViT (Fig.~\ref{fig:speed}). 

The main contributions of this work can be summarized into three aspects: 
\romannumeral 1) As a parallel to ViT, we propose a novel model with more efficiency, namely ViR, for image classification. 
\romannumeral 2) We design a nearly fully connected topology of the internal reservoir, with small-world characteristics, chaotic dynamics, and great memory capacity. 
Series ViR and parallel ViR are both developed, with the latter achieving better performance.
\romannumeral 3) ViR achieves superior performance on small-scale benchmarks than ViT, e.g., with about 15\% even 5\% the number of parameters of the ViT, and $20\% \sim 40\%$ the memory footprint of the ViT.

\section{Related Works}
\label{sec:formatting}
RC, mainly composed of Echo State Networks (ESNs) \cite{jaeger2001echo,jaeger2004harnessing} and Liquid State Machines (LSMs) \cite{maass2002real}, has been successively applied on speech recognition \cite{skowronski2006minimum}, time series prediction \cite{aswolinskiy2018time},  MIMO-OFDM \cite{mosleh2017brain}, smart grids \cite{hamedani2017reservoir}, etc. Physical RC \cite{tanaka2019recent} such as photonic RC \cite{van2017advances_Nanophotonics} is amenable to hardware implementation using a variety of physical systems, substrates, and devices.  However, few work have so far studied the behavior of such networks on computer vision. 

One of the related experiments has been carried out to classify digits of the MNIST dataset using a normal ESN \cite{schaetti2016echo}, which shows that RC can handle image data. More applications corresponding to images with RC are detailed in \cite{kleyko2017modality,jalalvand2018application,koprinkova2021reservoir}. Combining RC with other popular algorithms such as CNNs or Reinforcement Learning (RL) is a valid approach to enhance the performance of networks for image tasks. 
Tong and Tanaka \cite{tong2018reservoir} 
took full advantage of an untrained CNN to transform raw image data into a set of small feature maps as a preprocessing step of RC and achieved a high classification accuracy with a much smaller number of trainable parameters compared with Schaetti's work. Then, a novel practical approach, called RL with a convolutional reservoir computing (RCRC) model \cite{chang2020reinforcement} was proposed. A fixed random-weight CNN, used for extracting image feature maps, combined with an RC model, employed for extracting the time series features, is adopted for the evolution strategy of RL, which succeeded in decreasing computational cost to a large degree. However, such studies remain restricted because they simply treat RC as an auxiliary tool but not the core for image tasks. 

Based on the above issues, we propose an RC-based model specifically for image tasks, by exploring the internal topology of the reservoir. The proposed topology is originated from the Simple Cycle Reservoir (SCR), shown in \cite{rodan2010minimum}, which performs close to the traditional ESN. Later on, the updated one-Cycle Reservoirs with Regular Jumps (CRJ) \cite{rodan2012simple} has been proposed based on SCR, which obtains superior performances in most time-series tasks.

Recently, Verzelli \textit{et al.} \cite{verzelli2021input} made the best of the controllability matrix to explain the encoding mechanism and memory capacity of the reservoir with the topologies mentioned above, from the perspective of the mathematical characteristics of the matrix, i.e., the rank and the nullspace of the matrix. However, higher memory capacity usually does not correspond to higher precision prediction \cite{farkavs2016computational}, and the two are generally traded off in the "edge of chaos" state \cite{livi2017determination}.

Another recent related model is the deep RC. Taking ESN as an example, the typical architecture of deep ESN can be classified into the series ESN and the parallel ESN, deriving other deep structures \cite{liu2018analysis,gallicchio2017deep2,zhou2020deep}. 

There exist few studies combining RC with transformer \cite{vaswani2017attention}. Shen \textit{et al.} \cite{Shen2020ReservoirT} randomly initialized some of the layers in transformers without updates but obtained impressive performance in natural language processing. It only draws on the idea of initializing the parameters of the reservoir, not the "true" reservoir in essence.

\section{Methodology}

In the design of ViR, we first present the topology we utilized in the reservoir and show some formulations and characteristics to expound the working mechanism. Then, we describe the proposed ViR network and further present the case of the deep ViR. Finally, we give some aspects to analyze the intrinsic properties of the ViR.

\subsection{Nearly Fully Connected Topology}
\label{sub:topo}
Based on CRJ, we argue that for
an image classification task it is necessary to consider: 1) a simple fixed non-random reservoir topology--nearly full connection structure with self-loop connections; 2) the weight value 
$\textit{$r_i, r_j, r_k$} \in (0,1]$ for reservoir connections; 3) the signs of the weights in 2) are deterministically generated by a rule with a certain degree of randomness; 4) input neurons are randomly connected to the reservoir.

 \begin{figure}[htp]
   \centering
   \includegraphics[scale=0.5]{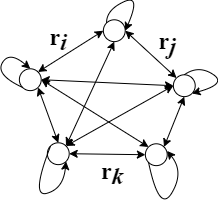}
   \caption{Nearly Fully Connected Topology. Neurons in the reservoir are nearly fully connected to others, containing three different types of weights.}
   \label{fig:hopfield}
   \vspace{-3mm}
 \end{figure}

Fig.~\ref{fig:hopfield} depicts the nearly fully connected topology. We first generate an $N \times N$ reservoir connection matrix \textbf{\textit{W}} matching the topology, in which all the weights have the same original value \textbf{$r_k$} and then we set:
 \begin{equation}
     \textbf{\textit{W}}_{1,\textit{N}}=\textbf{\textit{W}}_{\textit{q}+1,\textit{q}}=\textit{}{r_i} ,\qquad \textit{q}=1,...,\textit{N}-1
    \label{eq:Wq+1,q}
 \end{equation}
\textit{$r_j$} is the jump weight. Consider the jump size  $1<\ell<\textit{N}$. The first jump begins from neuron 1 to $1+\ell$, then from $1+\ell$ to $2\ell+1$, until $n\ell+1>N$, where \textit{n} is the number of jump. These bi-directional jump weights are reset to:
 \begin{equation}
 \begin{aligned}
     &\textbf{\textit{W}}_{1,\ell+1}=\textbf{\textit{W}}_{\ell+1,2\ell+1}=...\\
     &=\textbf{\textit{W}}_{\ell+1,1}=\textbf{\textit{W}}_{2\ell+1,\ell+1}=...=\textit{$r_j$}
   \end{aligned}   
  \label{eq:rj}
 \end{equation}
Finally, we randomly set an element, with the symmetrical element, to 0, indicating that the two neurons are disconnected with each other. Hence, \textbf{\textit{W}} mainly consists of \textit{$r_k$},  and the amount of \textit{$r_j,r_c$} is relatively small, with two 0 elements. 

For the randomness, there exists a random matrix with elements \textit{e} randomly generated in (0,1] with the same size as \textbf{\textit{W}}. If \textit{e} $<$ 0.5 (the threshold we set), then the sign of the weight in \textbf{\textit{W}}, corresponding to the same position as \textit{e}, will be $-$, otherwise $+$.
The input weight matrix \textbf{\textit{V}}, compared to \textbf{\textit{W}}, is a sparse matrix of the given shape with uniformly distributed values.

For the stability of the reservoir, \textbf{\textit{W}} is typically scaled as $\textbf{\textit{W}}\gets\alpha\textbf{\textit{W}} /|\lambda_{max}|$, where $|\lambda_{max}|$ is the spectral radius of \textbf{\textit{W}} and $0<\alpha<1$ is a scaling parameter \cite{jaeger2002tutorial}.

For the training process, assuming that $\textit{\textbf{u}(t)}=(u_1 (t), ..., u_K (t))$ means the activation of $K$ input units at time step $t$ from flattened patches, and $\textit{\textbf{x}(t)}=(x_1 (t), ..., x_M (t))$ represents the $M$ internal units in the reservoir with the output $\textit{\textbf{y}(t)}=(y_1 (t), ..., y_Q (t))$ containing $Q$ output units. The $M \times K$ matrix $\textbf{\textit{V}}$ stores input weights. The $M \times (K+M+Q)$ matrix $\textbf{\textit{W}}_{out}$ shows the connections from a reservoir to output units. Additionally, all the values mentioned above are real-valued. Apart from \textit{$\textbf{W}_{out}$}, all the matrices are not trainable and their entries are initialized with a suitable distribution and then fixed. Fig.~\ref{fig:vir core} shows the described structure.

The internal units are updated according to:
  \begin{equation}
  \textbf{\textit{x}}(\textit{t}+1)=\textit{f }  (\textbf{\textit{Vu}}(\textit{t}+1)+\textbf{\textit{Wx}}(\textit{t})+\textbf{\textit{b}}(\textit{t}+1))
  \label{eq:x(t+1)}
  \end{equation}
where \textit{f} is the activation function (typically sigmoidal); \textbf{\textit{b}}(\textit{t}+1) is an optional uniform i.i.d. noise. The readout is computed by:
  \begin{equation}
  \begin{aligned}
  \textbf{\textit{y}}(\textit{t}+1)=\textbf{\textit{W}}_{out}&[\textbf{\textit{u}}(\textit{t}+1); \textbf{\textit{x}}(\textit{t}+1); \textbf{\textit{y}}(\textit{t});\\ 
  &\textbf{\textit{u}}(\textit{t}+1)^2; \textbf{\textit{x}}(\textit{t}+1)^2; \textbf{\textit{y}}(\textit{t})^2]
  \end{aligned}
  \label{eq:yt+1}
  \end{equation}
where ';' is a concatenation operation.

As shown in Eq. \ref{eq:yt+1}, the essence of network is to approximate output weights $\textit{\textbf{W}}_{out}$ through training samples, to obtain the predictive ability of certain tasks. Hence, while training, it is essential to collect and store the state of the reservoir \textit{\textbf{x}}(\textit{t}) and the corresponding outputs \textit{\textbf{y}}(\textit{t})  in a matrix \textit{\textbf{X}} and \textit{\textbf{Y}} after warm-up \cite{jaeger2002tutorial}. The classic calculation method is the Least Square Method (LSM) \cite{lukovsevivcius2012reservoir}:
 \begin{equation}
     \textit{\textbf{W}}_{out}=(\textit{\textbf{X}}^T\textit{\textbf{X}})^{-1}\textit{\textbf{X}}^T\textit{\textbf{Y}}
     \label{eq:lsmwout}
 \end{equation}
where $(\ast)^T$ and $(\ast)^{-1}$ mean the transpose and the inverse of matrix operator, respectively. Usually, in case \textit{\textbf{X}} is ill-posed and irreversible, the ridge regression method \cite{shi2007support} is utilized (Eq.~\ref{eq:Ridge}) instead of LSM. For other optimizing algorithms refer to \cite{min2010multivariate, ozturk2005computing}.   
 \begin{equation}
     \textit{\textbf{W}}_{out}=(\textit{k}\textit{\textbf{I}}+\textit{\textbf{X}}^T\textit{\textbf{X}})^{-1}\textit{\textbf{X}}^T\textit{\textbf{Y}}
     \label{eq:Ridge}
 \end{equation}
where \textit{k} is a relatively small positive number called regular coefficient, and \textit{\textbf{I}} is an identity matrix. In our work, we use a trainable linear layer (LL) to approximate $\textit{\textbf{W}}_{out}$ and Eq. \ref{eq:yt+1} can be rewritten as:
  \begin{equation}
  \begin{aligned}
  \textbf{\textit{y}}(\textit{t}+1)=\textbf{\textit{LL}}&[\textbf{\textit{u}}(\textit{t}+1); \textbf{\textit{x}}(\textit{t}+1); \textbf{\textit{y}}(\textit{t});\\ 
  &\textbf{\textit{u}}(\textit{t}+1)^2; \textbf{\textit{x}}(\textit{t}+1)^2; \textbf{\textit{y}}(\textit{t})^2]
  \end{aligned}
  \label{eq:newyt+1}
 \end{equation}

\subsection{The Vision Reservoir (ViR)}

Fig.~\ref{fig:vir core} depicts the proposed model for image classification, and the key component is the ViR core which is composed of a reservoir with the internal topology mentioned above and a residual block.

The typical reservoir receives a time-series sequence as input. Similar to the treatment in ViT, some pretreatments are used to convert raw images to patches as the time series input. We reshape the original image $\textbf{x} \in \mathbb{R}^{\textit{H}\times\textit{W}\times\textit{C}}$ into a series of patches $\textbf{x}_\textit{p} \in \mathbb{R}^{\textit{N}\times(\textit{P}^2\cdot\textit{C})}$, where $(\textit{H, W})$ and $(\textit{P, P})$ respectively represent the resolution of \textbf{x} and $\textbf{x}_\textit{p}$, and $\textit{N} = \textit{H}\textit{W}/\textit{P}^2$ is the number of patches, also serving as the input time steps. With a trainable linear projection \textit{\textbf{E}} (Eq.~\ref{eq:input}), we flatten the patches and map to \textit{D} dimensions to make the inputs suitable for the reservoir. We refer to the output of this projection as the patch coding. Therefore, the input at time step \textit{t+n} can be written as:
 \begin{equation}
   \textbf{\textit{u}}(\textit{t+n}) = \textbf{x}^n_\textit{p}\textbf{\textit{E}}, \quad \textbf{\textit{E}} \in \mathbb{R}^{(\textit{P}^2\cdot\textit{C})\times\textit{D}}, \textit{n} = 0, 1,...,\textit{N}-1
   \label{eq:input}
  \end{equation}  

Then the ViR core receives the data. It consists of the nearly fully connected reservoir (Fig.~\ref{fig:hopfield}) and a residual block including a layer-norm (\textit{LN}) operation and a feed-forward (\textit{FF}) layer. The non-linear function in \textit{FF} is a Gaussian Error Linear Unit (GELU) \cite{hendrycks2016gaussian}. The processing of the reservoir has been shown in Eq.~\ref{eq:x(t+1)} and  Eq.~\ref{eq:yt+1}, and \textit{LN} is applied to  $\textbf{y}_\textit{r}$ (the output of the reservoir) with the feed-forward layer with a residual connection between them, shown as:
 \begin{equation}
   \textbf{\textit{y}}_\textit{c} = \textbf{\textit{FF}}(\textbf{\textit{LN}}(\textbf{\textit{y}}_\textit{r})) + \textbf{y}_\textit{r}
   \label{eq:residual}
  \end{equation}

A classification operation is attached to $\textbf{\textit{y}}_\textit{c}$ (the output of the ViR core ), which is implemented by a MLP, and we can obtain the image representation and the final classification results \textbf{y} by Eq. \ref{eq:y}:
 \begin{equation}
   \textbf{\textit{y}} = \textbf{\textit{LN}}(\textbf{\textit{y}}_\textit{c})
   \label{eq:y}
  \end{equation}
  
\subsection{The Deep ViR}

We further stack reservoirs to get the deep ViR to enhance the network performance.

\begin{figure}[htp]
    \centering
    \includegraphics[scale=0.21]{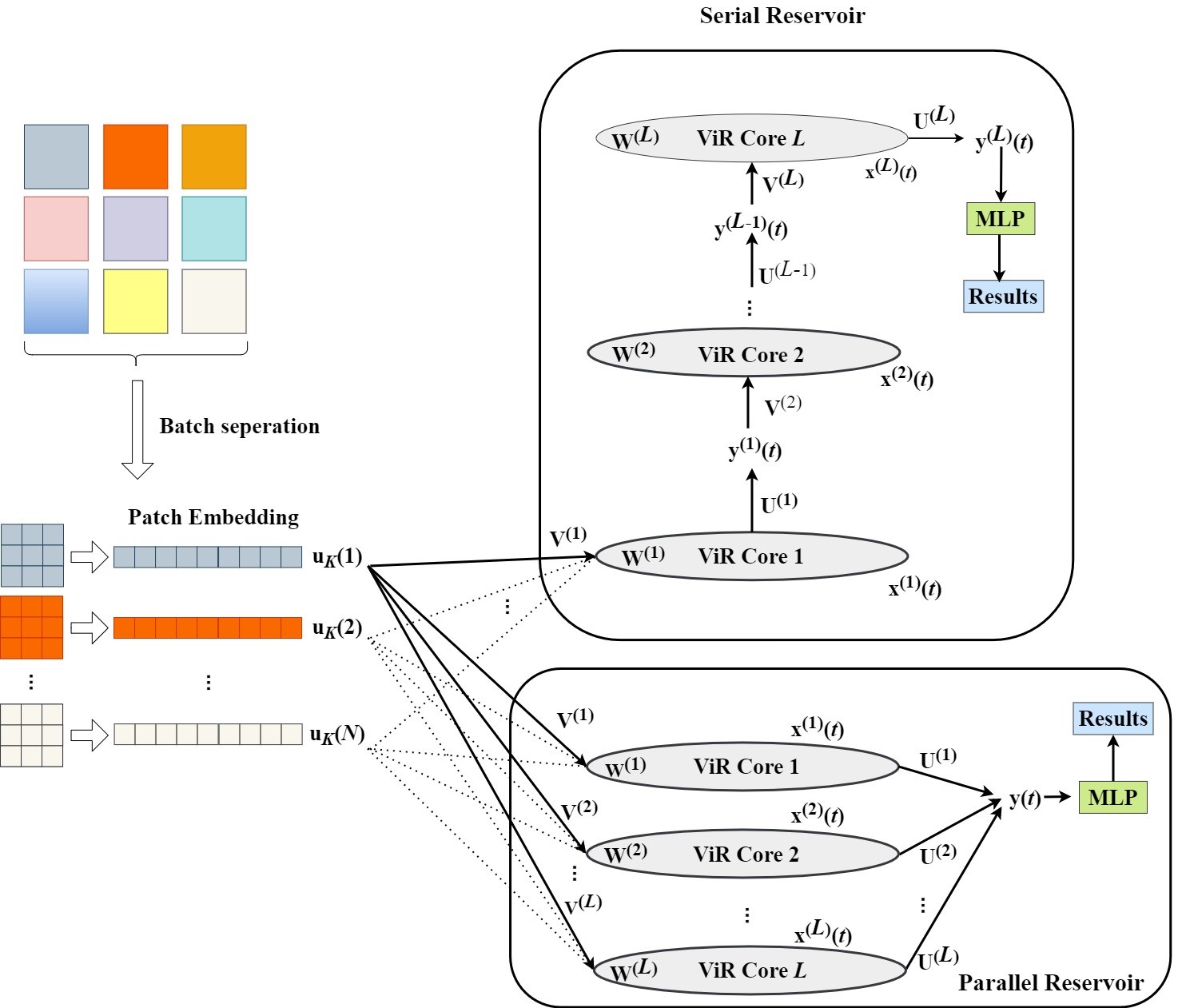}
    \caption{The structure of the deep ViR. The upper one is the series reservoir, and the lower one refers to the parallel reservoir.}
    \label{fig:sd}
    \vspace{-3mm}
  \end{figure}

The first one is the series reservoir consisting of \textit{L} reservoirs depicted in Fig.~\ref{fig:sd}. Similarly, the input matrices  $\textbf{\textit{\textit{V}}}^{(l)}$ and reservoir connection matrices $\textbf{\textit{W}}^{(l)}$ are constant and generated as mentioned above, \textit{l} $\in$ $\{$1,2,...,\textit{L}$\}$, with similar training process described in subsection \ref{sub:topo}. 

$\textbf{\textit{U}}^{(l)}$ is the output matrix. Units in reservoir \textit{l} updating and the output of reservoir \textit{l} are given by:
  \begin{equation}
  \textit{$\textbf{x}^{(l)}$} (\textit{t}+1)=\textit{f } (\textit{$\textbf{V}^{(l)}$} \textit{$\textbf{y}^{(l-1)}$} (\textit{t}+1)+\textit{$\textbf{W}^{(l)} \textbf{x}^{(l)}$} (\textit{t}))
  \end{equation}  
  \begin{equation}
  \textit{$\textbf{y}^{(l)}$} (\textit{t}+1)=\textit{$\textbf{U}^{(l)}$}[\textit{$\textbf{y}^{(l)}$} (\textit{t}); \textit{$\textbf{x}^{(l)}$} (\textit{t}+1); \textit{$\textbf{y}^{(l)}$} (\textit{t})^2; \textit{$\textbf{x}^{(l)}$} (\textit{t}+1)^2]
  \end{equation}
where \textit{l} $\in$ $\{$1,2,...,\textit{L}$\}$ and $\textbf{\textit{y}}^{(0)} (\textit{t}) = \textbf{\textit{u}}(\textit{t})$, $\textbf{\textit{y}}^{(L)} (\textit{t}) = \textbf{\textit{y}}(\textit{t})$.

The other one is a parallel reservoir, depicted in Fig.~\ref{fig:sd}.

\textit{L} reservoirs simultaneously receive the same input sequence \textit{\textbf{u}(t)} and all \textit{L} reservoirs can be trained simultaneously referring to the above process. The units in \textit{L} reservoirs and the output units are updated as follows:   
\begin{equation}
  \textit{$\textbf{x}^{(l)}$} (\textit{t}+1)=\textit{f } (\textit{$\textbf{V}^{(l)}$} \textbf{\textit{u}} (\textit{t}+1)+\textit{$\textbf{W}^{(l)} \textbf{x}^{(l)}$} (\textit{t}))
  \end{equation}  
  \begin{equation}
  \textit{$\textbf{y}^{(l)}$} (\textit{t}+1)=\textit{$\textbf{U}^{(l)}$}[\textit{$\textbf{y}^{(l)}$} (\textit{t}); \textit{$\textbf{x}^{(l)}$} (\textit{t}+1); \textit{$\textbf{y}^{(l)}$} (\textit{t})^2; \textit{$\textbf{x}^{(l)}$} (\textit{t}+1)^2]
  \end{equation}
 The final output of a parallel reservoir is the arithmetic mean of \textit{L} reservoir outputs, given by:
   \begin{equation}
       \textit{\textbf{y}} (\textit{t})=\sum_{l=1}^L \textit{$\textbf{y}^{(l)}$} (\textit{t}) / \textit{L}
   \end{equation}

\subsection{Theoretical Analyses}
\label{theory}
Usually, we evaluate the quality of a reservoir through three indicators: the Small-World (SW) characteristics, the Largest Lyapunov Exponent (LLE), and the Memory Capacity (MC) \cite{nakajima2021reservoir_book}. The reservoir with SW characteristics has a stronger information processing ability and faster information dissemination speed than the common reservoir \cite{jaeger2004harnessing,albert2002statistical}. LLE of a reservoir represents the dynamics and stability, and usually the value of LLE should be approximately equal to 0 for great performance. MC is the property of a reservoir to keep the previous input information, related to the data processing capability.

\textbf{Small-World Characteristics:} RC is a brain-inspired neural network, which mimics the connections of cerebral neurons, cortical neural connectivity has been shown to exhibit a small-world (SW) network topology \cite{watts1998collective}, which has a shorter average path length and larger clustering coefficient than regular networks. In a reservoir, a small clustering degree means that dynamic information flow through the reservoir nodes is not ‘too cluttered’. Also, a small average path length can allow for the representation of a variety of dynamic time scales.

Due to the bi-directional connections, we can view the interconnection topology as an undirected graph \textbf{\textit{G}}=(\textbf{\textit{J, E}}), \textbf{\textit{J}} and \textbf{\textit{E}} representing neurons and the connections in the reservoir, respectively. This is similar to the kernel connections in the ViT. 
Define $\textbf{\textit{l}}_\textbf{\textit{G}}$ to be the average path length between vertex pairs in \textbf{\textit{J}}, which is calculated by \cite{newman2003structure}:
   \begin{equation}
      \textbf{\textit{l}}_\textbf{\textit{G}}=\frac{2}{M(M+1)} \sum_{\substack{i \geq j}} d_{ij}
      \label{lg1}
   \end{equation}
where $d_{ij}$ is the geodesic distance from vertex \textit{i} to vertex \textit{j}, and \textit{M} is the number of nodes (equal to the number of neurons in the reservoir). If $i=j$, the distance is 0 \cite{newman2003structure}. Moreover, if there are no self-loop connections, Eq. \ref{lg1} should be rewritten as:
   \begin{equation}
      \textbf{\textit{l}}_\textbf{\textit{G}}=\frac{2}{M(M-1)} \sum_{\substack{i > j}} d_{ij}
      \label{lg2}
   \end{equation}

Assuming that the neighborhood $|\textit{M}_i|$ of the node $\textit{v}_i$ is the nodes next to $\textit{v}_i$, $\textit{M}_i$ is the number of nodes in $|\textit{M}_i|$. 
The local clustering coefficient $\textbf{\textit{C}}_\textbf{\textit{i}}$ of $\textit{v}_i$ is given by a ratio, i.e., the connecting edges 
between nodes in the neighborhood divided by the number of possible connecting edges between them, shown in Eq. \ref{eq:ci} \cite{watts1998collective}:
   \begin{equation}
      \textbf{\textit{M}}_\textbf{\textit{i}}={\textit{v}_j:\textit{e}_{ij} \in \textbf{\textit{E}} \vee \textit{e}}_{ji} \in \textbf{\textit{E}}
   \end{equation}
      \begin{equation}
      \textbf{\textit{C}}_\textbf{\textit{i}}=\frac{2|{\textit{e}_{jk}:\textit{v}_j, \textit{v}_k \in \textbf{\textit{M}}_i,  \textit{e}}_{jk} \in \textbf{\textit{E}}|}{M(M-1)}
      \label{eq:ci}
   \end{equation}
The overall clustering level $\bar{\textit{C}}$ is
$      \bar{\textbf{\textit{C}}}=\frac{1}{M} \sum_{\substack{i=1}}^{\substack{M}} C_i$.

The degree of SW, named small-worldness, is given as:
      \begin{equation}
      \delta=\frac{C}{C_0}/\frac{l_G}{l_0}
   \end{equation}
where $\textit{C}_0$ and $\textit{l}_0$ represent the average path length and clustering level of a regular network with similar size. If $\delta > 1$, the network is a SW network \cite{humphries2006brainstem}.

\textbf{Lyapunov Exponent:} The Largest Lyapunov Exponent (LLE) \cite{wolf1985determining} represents the stability of reservoir dynamics, expounding the sensitivity of initial conditions of a system to small perturbations, defined as \cite{tsuruta2003small}:
\begin{equation}
    \textbf{$\lambda$}=\lim_{k\to\infty}\frac{1}{k}ln(\frac{\gamma_k}{\gamma_0})
\end{equation}
where $\gamma_0$ is the initial distance between the perturbed and the unperturbed trajectory, $\gamma_k$ is the distance at time k. For ordered dynamic systems, $\lambda < 0$ and for chaotic systems $\lambda > 0$. 
At $\lambda \approx 0$, a phase transition occurs (called the critical point, or edge of chaos) with the best computational capability \cite{bertschinger2004real,boedecker2012information}.

\textbf{Memory Capacity:} Memory Capacity (MC) \cite{jaeger2001short} is another metric to measure the learning performance of our model. With receiving a
random input \textbf{u}(t) at a time \textit{t}, the reservoir is trained to generate the desired output $\textbf{y}_{d}(\textit{t})=u(t-\tau)$.  The output \textbf{y} is learned from \textbf{u}(t) which was $\tau$ steps earlier. Then the MC is given as:
\begin{equation}
    \textbf{MC}_{\tau}=\frac{cov^2 (\textbf{u}_{\tau},\textbf{y})}{v^{2}(\textbf{u}_{\tau})v^{2}(\textbf{y})}, \qquad \textbf{MC} = \sum_{\tau=1}^T \textbf{MC}_{\tau}
\end{equation}
where $cov^2(\textbf{u}_{\tau},\textbf{y})$ indicates the covariance between the true value $\textbf{u}_{\tau}$ and the  predicted value \textbf{y}. $v^{2}(\ast)$
means the variance of a series $\ast$. T is the maximal time delay we set.

We have experimentally demonstrated the SW characteristics, largest Lyapunov exponents, and memory capacities of our model, in Section \ref{041}.

\section{Experiment Results}
\label{04}
 The comparative study between the proposed ViR and the common ViT model is carried out on three classical datasets, 
 i.e. MNIST, CIFAR10, and CIFAR100\cite{2012Learning}.  
 We also compare the number of parameters in the ViR with the ViT and analyze the convergence speed as well as the memory footprint of our models. Further, the robustness is tested on CIFAR10-C \cite{hendrycks2019benchmarking}. 

In this section, original ViT is named ViT-Base \cite{dosovitskiy2020image} with several changes, as shown in Table \ref{tab:setup}.

  \begin{table}[htbp]
    \centering
          \caption{The system parameters of the ViR and the ViT. \textit{N} is the number of neurons in  a reservoir, and $\alpha$ is a scaling parameter of the spectral radius of \textbf{\textit{W}}; \textit{SD} is the sparse degree of input matrix \textbf{\textit{V}}. And \textit{$r_i, r_j, r_k$} and jump size are detailed in subsection \ref{sub:topo}. In the ViT row, patch size is the same for all tested datasets.}
      \begin{tabular}{c|c|c}
            \hline & 
            \multicolumn{1}{|c}{\textbf{parameters}} &
            \multicolumn{1}{|c}{\textbf{values}} 
            \\ \hline
            \multirow{2}*{ViR}
        & (\textit{N}, $\alpha$, \textit{SD})& (1000, 0.9, 0.05)   \\
      & (\textit{$r_i, r_j, r_k$}, jump size) &   (0.05, 0.5, 0.08, 137)\\
      \hline
       \multirow{4}*{ViT}
      & Hidden size& 512\\
      &MLP size&2048  \\
      &Heads &12\\
      & Patch size &$4\times4$\\
      \hline
      \end{tabular}
    \label{tab:setup}
  \end{table}

\subsection{Small-Worldness, Lyapunov Exponent and Memory Capacity}
\label{041}
Table \ref{tab:sw} shows the normalized average path length  $\frac{l_G}{l_0}$, clustering coefficient $\frac{C}{C_0}$, and small-worldness $\delta$, $\frac{C}{C_0}/\frac{l_G}{l_0}$. The baseline is the classic regular network and random network detailed in  \cite{kawai2019small}.

  \begin{table}[htbp]
    \centering
    \small
    \caption{Small-worldness of the reservoirs with 1000 neurons. $l_0$ and $C_0$ are the average path length and clustering level of the regular network, correspondingly. Both are usually equal to 1.}
      \begin{tabular}{c|c|c|c}
            \hline & 
            \multicolumn{1}{|c|}{\textbf{$l_G/l_0$}} &
            \multicolumn{1}{|c|}{\textbf{$C/C_0$}} &
            \multicolumn{1}{c}{\textbf{$\delta$}}
            \\ \hline
      Regular Network &  1.00  & 1.00  & 1.00  \\
      Ours &  0.98  &  0.99 & 1.02 \\
     Random Network &  0.09 &  0.02  & 0.20 \\
      \hline
      \end{tabular}
    \label{tab:sw}
    \vspace{-3mm}
  \end{table}
Our model has a lower clustering level and a smaller average path than the regular one. It still exhibits the SW property, 
referring to the fact that our topology has great potential for information calculation and processing. Similar to the brain \cite{rubinov2015wiring}, a small average path indicates low wiring cost for physical implementations. By applying a standard algorithm given in \cite{shimada1979numerical}, inputs are sampled from Gaussian distribution \cite{verstraeten2010memory} for the memory capacity and the largest Lyapunov exponents. 

\begin{figure}[htbp]
 \centering
	\begin{minipage}[t]{0.45\textwidth}
	\centering          
	\includegraphics[scale=0.32]{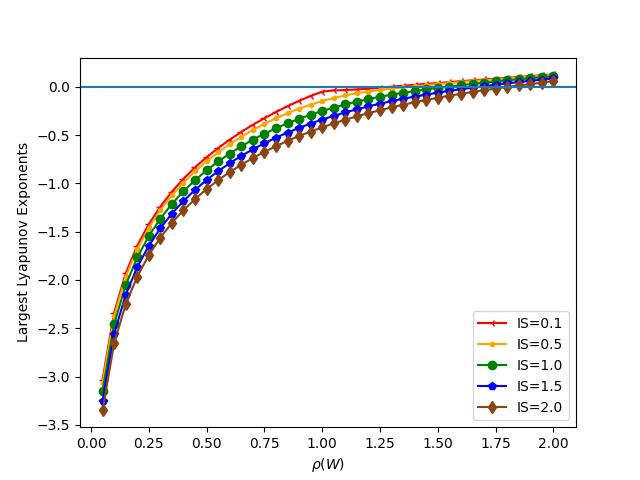}   
	\vspace{-2mm}
	\caption*{(a). Lyapunov exponents versus spectral radius.}
	\end{minipage}
	\begin{minipage}[t]{0.45\textwidth}
	\centering   
	\includegraphics[scale=0.32]{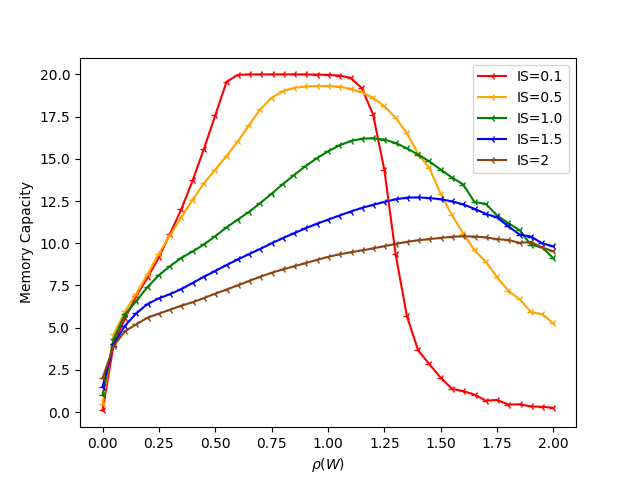}  
	\vspace{-2mm}
	\caption*{(b). Memory capacities versus spectral radius.}
	\end{minipage}
 \caption{Lyapunov exponents and memory capacities. 
 The horizontal axe denotes the spectral radius of \textbf{W}. The vertical axe denotes the Lyapunov exponents and memory capacities in (a) and (b), respectively. IS is a scaling factor to the input weight matrix $\textbf{W}_{in}$.}
 \label{fig:lymc}
 \vspace{-3mm}
 \end{figure}

Fig.~\ref{fig:lymc}(a) shows we get the peak Memory Capacity (MC)  when the spectral radius \textbf{$\rho$}(W) is around 1. Also, Fig.~\ref{fig:lymc}(b) shows that the largest Lyapunov Exponent (LLE, defined in subsection \ref{theory}) is close to 0 when \textbf{$\rho$}(W) is around 1$\sim$1.5. Meanwhile, the influences of the input scaling (IS) to MC and LLE are also shown in Fig.~\ref{fig:lymc}. And the results indicate that we should constrain the \textbf{$\rho$}(W) and the IS to about 1 for image tasks.

\subsection{Main Results with Training from Scratch}

  \begin{table*}[htbp]
    \centering
    \caption{Comparisons between ViR and the ViT models without pre-training on prevalent image classification datasets. The digital suffix means the number of the ViR layers or encoders of the ViT. "M" is expressed as a unit symbol of one million.
    }
      \begin{tabular}{c|c|c|c|c|c|c}
            \hline&
            \multicolumn{2}{|c|}{\textbf{MNIST}} &
            \multicolumn{2}{|c|}{\textbf{CIFAR10}} &
            \multicolumn{2}{|c}{\textbf{CIFAR100}}
             \\ 
   \cline{2-7}
    \multirow{2}{*}{}&accuracy[\%] & parameters[M] &accuracy[\%] & parameters[M] &accuracy[\%] & parameters[M]\\
    \hline
      ViT-1 &  98.31 &3.48 & 64.81 &3.49 & 35.21  &3.53\\
      ViR-1 &  98.92  &\textbf{0.60}&  78.03 &\textbf{0.61}& 51.41 &\textbf{0.62} \\
      ViT-3 &  98.49  &10.42&  74.78  &10.44& 43.63 & 10.47\\
      Series-ViR-3 & 98.27 &1.79&  73.39 &1.80& 46.91 & 1.81\\
      Parallel-ViR-3 &   98.95  &1.59&  78.51  &1.60& 52.25 & 1.61\\
      ViT-6 & 98.54   & 20.84  & 77.68 & 20.85 & 44.14 &20.88\\
      Series-ViR-6 & 98.45 &  3.59 & 74.3 & 3.60 &46.87 &3.61\\
      Parallel-ViR-6 &  99.05  &3.19&  79.13  &3.20  &  51.73&3.21\\
      ViT-12 & 98.72  &41.66&  79.72  &41.67&  45.61 & 41.71 \\
      Series-ViR-12& 98.68 &7.17&  75.63  &7.18& 47.02 & 7.19\\
      Parallel-ViR-12& \textbf{99.17} &6.24& \textbf{80.02} &6.25& \textbf{51.98} & 6.26\\ 
      \hline
      \end{tabular}

    \label{tab:accuracy}
    \vspace{-3mm}
  \end{table*}

Without any pre-training, we compare our models ViR-1, ViR-3, ViR-6, and ViR-12 with ViT-1, ViT-3, ViT-6, and ViT-12 by performing image classification tasks on MNIST, CIFAR10, and CIFAR100. 
Table \ref{tab:accuracy} shows the classification accuracy and the number of parameters. 
It can be seen that ViR-1 has more competitive results than ViT-12 and therefore ViR models have a promising advantage in saving parameters.
According to the aforementioned analysis, the reservoir with a nearly fully connected topology has great information processing capability and rich dynamics. Hence, our model is suitable for handling image tasks.

\textbf{Accuracy:} All the ViR models except Series-ViR perform better than ViT models with the same depths. Depths are the layers of the ViR or the encoders of the ViT. Interestingly, the shallow model ViR-1 usually rivals ViT-1, ViT-3, and ViT-12. This indicates that shallow ViR models have a competitive performance compared with the deep ViT models, but with lower time costs and fewer parameters. The great MC of the reservoir donates the high accuracy to some degree.

\textbf{Convergence Speed:} As depicted in Fig.~\ref{fig:speed}, the initial test accuracy of the ViR is already close to the best results of the ViT. Meanwhile, the ViT-12 suffers over-fitting. The convergence speed of most ViR models is much faster than ViT. This reflects the SW characteristics and rich dynamics from LLE. 

\textbf{The Number of Parameters:} Due to the fact that training a reservoir is limited to the output layer, $\textbf{\textit{W}}_{out}$, much fewer parameters are trained compared to other models. This is one of the attractive characteristics of RC. In this aspect, the ViR has the promising advantage compared with ViT.

Table \ref{tab:accuracy} shows the trainable parameters of the ViR and the ViT on different datasets with different layers. The number of parameters of the ViR is about 15\% of the ViT with the same depth. But the comparison between shallow ViR models and deep ViT models shows that the number of parameters of the ViR could be about 5\% of the ViT while guaranteeing an acceptable accuracy. 

Table \ref{tab:accuracy}  also indicates that, with the same depth, the series reservoir usually performs worse than the parallel one. One possible explanation is that: the series one always randomizes the input to the next reservoir, having no substantial help for tasks. However, the parallel one randomizes the inputs only once and then gets the arithmetic mean results. As a result, the enhancement of accuracy is predictable. With the increase of reservoir layers, the gains from deep structures can't offset the influence of time and computational costs compared to shallow ones.
This reveals that it is not necessary to stack too many reservoirs.

\subsection{Memory Footprint}
\begin{figure}[htbp]
 \begin{center}
  \includegraphics[scale=0.58]{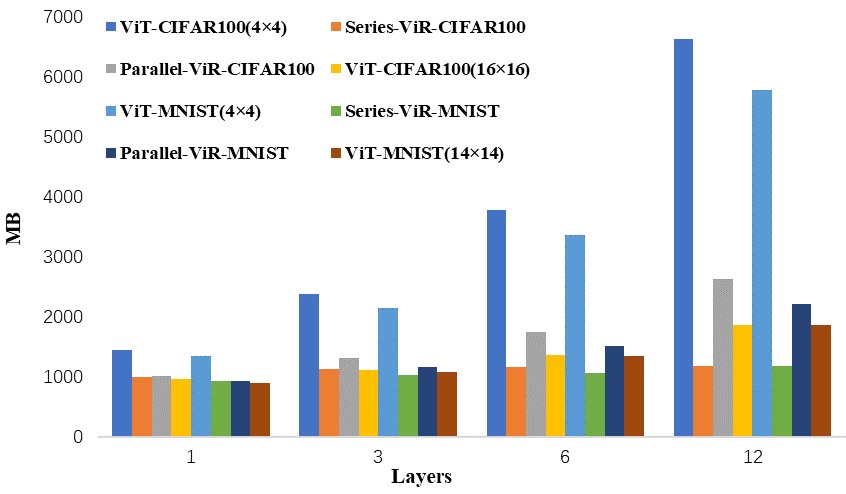}
  \end{center}
  \vspace{-2mm}
  \caption{Comparisons of memory footprint on MNIST and CIFAR100 datasets with the patch sizes of
  $4\times4, 14\times14$, and $16\times16$. 
  }
 \label{fig:mf}
 \vspace{-3mm}
 \end{figure}

Considering the potential crash risk from memory overloading, it is necessary to evaluate and reduce its usage in a single task while ensuring acceptable results. The memory footprint of the ViR, compared with ViT models, on MNIST and CIFAR100 datasets, is shown in Fig.~\ref{fig:mf}.
For small patch size, we observed that the memory footprint of the ViT models is about 2.5 to 5 times to ours, with the same depth. 
Although the large patch size of the ViT could reduce the memory footprint, our models (with small patch size) still have a slight superiority in such cases. 
It is worth noticing that, for both the series ViR and the parallel ViR the memory footprints within the optimal number of layers are nearly equal. 

  \begin{table*}[htbp]
    \centering
    \caption{Robustness for evaluating the influence of corruptions of input images. The $\textbf{CE}_c$ value for a certain type of corruption is shown from column $2$ to $9$, correspondingly. The \textbf{CE} value is the arithmetic mean corruption error of the corruptions in Noise, Blur, Weather, and Digital columns. In these cases, models are trained only on clean CIFAR10 images and tested on CIFAR10-C.}
      \begin{tabular}{c|c|c|c|c|c|c|c|c|c}
            \hline & \multicolumn{2}{|c|}{\textbf{Noise}} &
            \multicolumn{2}{|c|}{\textbf{Blur}} &
            \multicolumn{2}{|c|}{\textbf{Weather}} &
            \multicolumn{2}{|c|}{\textbf{Digital}} &
            \multicolumn{1}{|c}{\multirow{2}{*}{\textbf{CE}}}
            \\
            \cline{2-9}
       \multirow{2}{*}{} & Gaussian & Shot & Defocus & Motion & Snow & Frost& Contrast & JPEG &\\
       \hline
      ViT-1 & 1.8471&1.9345 & 1.8551&1.8546 &1.9271 &1.9468 &1.9635 &1.9736 &1.9128  \\
      ViT-3 & 1.7218&1.7263 &1.6837 & 1.7728& 1.7978& 1.7761&1.7726 &1.7379 &  1.7486\\
      ViT-6 & 1.6662&1.7562 & 1.6608& 1.6907& 1.6845&1.6973 &1.7612 &1.7261 & 1.7054 \\
      ViT-12 &1.5557&1.7023 &1.5234 &1.5343 & 1.5310&1.6287 & 1.6215& 1.6779& \textbf{1.5969}\\
      ViR-1 & 1.6441 & 1.8079& 1.6573& 1.7385& 1.7294 &1.8717 &1.7629 & 1.7506&1.7447  \\
      Series-ViR-3 & 1.8835&1.9225  &1.7164 &1.8763 & 2.0488 &2.2069 &2.1572 & 1.8542&1.9582 \\
      Series-ViR-6 &1.9820 &2.2625 &2.0411 &2.1119 & 2.4030 &2.4629 &2.3488 &2.3164 & 2.2411\\
      Series-ViR-12& 1.8219& 2.4180&2.3383 &2.3281 &2.8341 &2.8164 &2.8712 & 2.4978&2.4907 \\
      Parallel-ViR-3 &1.6365 &1.7849 &1.6159 &1.7034 & 1.7152 &1.8102 &1.7603 &1.8061 &1.7291 \\
      Parallel-ViR-6 &1.5035 & 1.8193&1.6361 &1.6801 &1.7296 &1.8739 &1.7829 &1.6950 & 1.7151\\
      Parallel-ViR-12&1.4911 & 1.8188& 1.6006&1.6858 & 1.7239& 1.8758& 1.7242& 1.6739& \textbf{1.6993}\\ 
      \hline
      \end{tabular}
    
    \label{tab:robust}
    \vspace{-3mm}
  \end{table*}

\subsection{Robustness}
We evaluate the robustness of the ViR from two aspects: i.e. corruptions of input images and perturbations of system hyperparameters. 

\textbf{Corruptions of input images}. The input images are selected from the CIFAR10-C dataset, which consists of imposed corruptions from noise, blur, weather and digital categories on the original CIFAR10 dataset \cite{hendrycks2019benchmarking}.

To achieve the comprehensive evaluation of the robustness to a given type of corruption, we score the classification performance across five corruption severity levels, denoted by \textit{s} ($1 \le s \le 5$), and aggregate these scores according to: 
   \begin{equation}
      \textbf{CE}_c=\sum_{s=1}^5 \textbf{E}_{s,c}-\textbf{E}_{clean}, \textbf{CE}=\frac{\textbf{CE}_c}{n}
   \end{equation}
where $\textbf{E}_{clean}$ is the top-1 error rate on clean CIFAR10 dataset; $\textbf{E}_{s,c}$ is the top-1 error rate on CIFAR10-C, \textit{c} is the type of corruptions. \textbf{CE} is arithmetic mean of all types of $\textbf{CE}_c$, and \textit{n} represents the number of corruption types. It should be pointed out that, the smaller value of $\textbf{CE}_c$ or $\textbf{CE}$ represents the stronger robustness.

Table \ref{tab:robust} compares the robustness performance
by means of $\textbf{CE}_c$ and $\textbf{CE}$. 
The results show that the number of layers within parallel-ViR has little influence on robustness. 
With the increasing encoding numbers of the ViT, it has increasingly robustness, which is contrary to in the cases of the series ViR model. 

 \begin{figure}[htp]
 \centering
     \vspace{-3mm}
	\begin{minipage}[t]{0.45\textwidth}
	\centering          
	\includegraphics[scale=0.25]{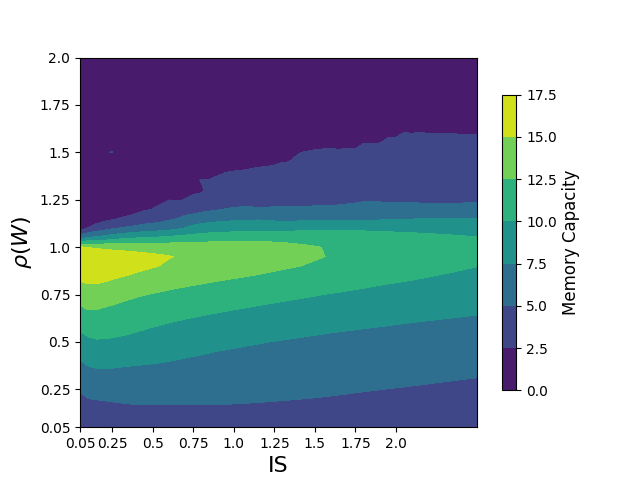} 
	 \vspace{-3mm}
	\caption*{(a).MC}
	\label{fig:mcheat}
	\end{minipage}

	\begin{minipage}[t]{0.45\textwidth}
	\centering   
	\includegraphics[scale=0.25]{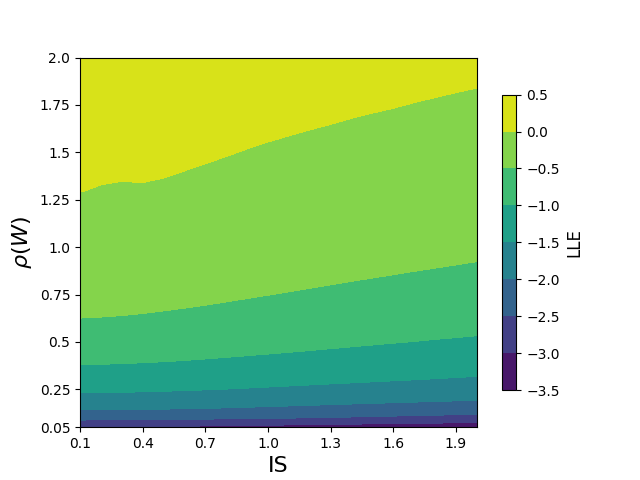} 
	 \vspace{-3mm}
	\caption*{(b).LLE}
	\label{fig:lleheat}
	\end{minipage}
 \caption{The influence of Input Scaling (IS). Considering the Memory Capacity (MC) (a) and the Largest Lyapunov Exponent (LLE) (b), the value of the spectral radius and IS should be close to 1 when training.}
 \label{fig:heat}
 \vspace{-3mm}
\end{figure}

\textbf{Perturbations of system hyperparameters}. 
From Fig.~\ref{fig:speed}, we can see that the number of layers has a smaller impact on our models, compared with the ViT, perhaps because the spectral radius of \textbf{\textit{W}} in the ViR is constrained to be around 1 when training \cite{nakajima2021reservoir_book}.
Moreover, the influence of IS is also reduced because of the constrained spectral radius, shown in Fig.~\ref{fig:heat}.
It has the same conclusion for other hyperparameters of the ViR, such as the influence of the number of neurons \textbf{\textit{N}}, \textit{$r_i, r_j, r_k$} and the jump size. SW characteristics support our experimental results, which make RC have strong robustness.
However, the hyperparameters in ViT, such as the number of heads, patch size, and MLP size etc., will significantly impact the classification accuracy. This indicates that our model has a certain degree of robustness to the perturbations of system hyperparameters.

\section{Conclusion}
In this paper, we proposed a novel model, namely ViR, for image classification. 
Similar to the ViT, the ViR treats the image as a sequence of pixel patches. 
The sequences are then processed by using a reservoir with a nearly fully connected topology. 
This simple and low-cost network model performs well, compared with the ViT models, on MNIST, CIFAR10, and CIFAR100. 
The ViR has small-world characteristics, chaos dynamics, and great memory capacities.
The ViR exceeds the ViT models, with the learning from scratch, on image classification, with a lower memory footprint, a smaller number of parameters, and faster training speed.
Encouraging by these experimental results, further study will be focused on the following three aspects, 
\romannumeral 1) theoretical explanations for the working mechanism of the reservoir,
\romannumeral 2) pre-training on large-scale datasets,
\romannumeral 3) applying the ViR to more computer vision tasks.

{\small
\bibliographystyle{ieee_fullname}
\bibliography{egbib}
}

\end{document}